\documentclass{article}

\usepackage[preprint]{neurips_2021}

\usepackage[utf8]{inputenc} 
\usepackage[T1]{fontenc}    
\usepackage{hyperref}       
\usepackage{url}            
\usepackage{booktabs}       
\usepackage{amsfonts}       
\usepackage{nicefrac}       
\usepackage{microtype}      
\usepackage{xcolor}         
\usepackage{mathtools}
\usepackage{graphicx}
\usepackage{subfigure}
\usepackage{booktabs} 
\usepackage{geometry}
\usepackage{tabularx}

\usepackage{amsmath}        
\setcitestyle{numbers, open={[}, close={]}}
\setcitestyle{square}

\DeclareMathOperator*{\argmax}{arg\,max}
\DeclareMathOperator*{\argmin}{arg\,min}

\newcommand{\bbR}{\mathbb{R}}
\newcommand{\bbN}{\mathbb{N}}
\newcommand{\bbE}{\mathbb{E}}
\newcommand{\calN}{\mathcal{N}}
\newcommand{\calD}{\mathcal{D}}
\newcommand{\GP}{\mathcal{GP}}
\newcommand{\bfx}{\textbf{x}}
\newcommand{\bfy}{\textbf{y}}
\newcommand{\bfk}{\textbf{k}}
\newcommand{\xstart}{\bfx^{\mathrm{start}}}
\newcommand{\xnext}{\bfx^{\mathrm{next}}}
\newcommand{\ystart}{y_{\mathrm{start}}}

\title{Differentiating Policies for Non-Myopic Bayesian Optimization}

\author{%
  Darian Nwankwo 
    \\
  Department of Computer Science\\
  Cornell University\\
  Ithaca, NY 14850 \\
  \texttt{don4@cornell.edu} \\
   \And
   David Bindel \\
   Department of Computer Science \\
   Cornell University \\
   Ithaca, NY 14850 \\
   \texttt{bindel@cornell.edu} \\
}

\begin{document}


\maketitle


\begin{abstract}
%
%

  Bayesian optimization (BO) methods choose sample points by optimizing an acquisition function derived from 
  a statistical model of the objective.  These acquisition functions are chosen to balance sampling regions with
  predicted good objective values against exploring regions where the objective is uncertain.
  Standard acquisition functions are myopic, considering only the impact of the next sample,
  but non-myopic acquisition functions may be more effective.
  In principle, one could model the sampling by a Markov decision process, and optimally choose the next sample
  by maximizing an expected reward computed by dynamic programming; however, this is infeasibly expensive.
  More practical approaches, such as rollout, consider a parametric family of sampling policies.
  In this paper, we show how to efficiently estimate rollout acquisition functions and their gradients,
  enabling stochastic gradient-based optimization of sampling policies.
\end{abstract}

\section{Introduction}

Bayesian optimization (BO) algorithms are sample-efficient methods for global optimization of
expensive continuous ``black-box'' functions for which we do not have derivative information.
BO builds a statistical model of the objective based on samples, and chooses new sample locations by
optimizing an acquisition function that reflects the value of candidate locations.
Most BO methods use {\em myopic} acquisition functions that only consider the impact of the next sample.
{\em Non-myopic} strategies~\cite{Yue2019} may produce high-quality solutions with
fewer evaluations than myopic strategies.
One can define an ``optimal'' non-myopic acquisition function in terms of the expected reward
for a Markov decision process.
But because of the curse of dimensionality, an exact dynamic programming approach to compute
such a function is computationally intractable in almost any problem of interest.
Approximate methods such as {\em rollout} are more practical, but remain expensive.

This paper aims to make non-myopic BO more practical. 
In particular, our main contributions are:
\begin{itemize}
    \item We compute rollout acquisition functions via quasi-Monte Carlo integration and use variance 
    reduction techniques to decrease the estimation error.
    \item We introduce a trajectory-based formulation for deriving non-myopic acquisition functions 
    using myopic functions—or any heuristic—as base heuristics.
    \item We show how to differentiate rollout acquisition functions given a differentiable base policy.
\end{itemize}
We show on a suite of test problems that we tend to do better than
myopic policies when considered the best decrease in objective function
from the first iteration to the last.

The subsequent sections of this paper discuss background and related 
work, including Gaussian Process Regression, Myopic Bayesian 
Optimization, Non-Myopic Bayesian Optimization, and Rollout 
Acquisition Functions (Sec. ~\ref{sec:background}). Sec. 
~\ref{sec:models} covers our models and methods, 
detailing our model and variance reduction techniques. We then 
explore our experimental results and their implications in Sec. 
~\ref{sec:experiments}. Limitations are discussed in 
Sec.\ref{sec:limitations}. The paper concludes with our final 
thoughts in Sec. ~\ref{sec:conclusion}.


\section{Background and related work}
\label{sec:background}

Non-myopic Bayesian optimization has received a lot of attention over the past 
few years \cite{Ginsbourger2010, Osborne2009, Ling2016, Lam2016, Lam2017, Gonzalez2016, Frazier2018, Frazier2019, Yue2019}.
Much of this work involves {\em rollout} methods, in which one considers the expected best value seen over
$h$ steps of a standard (myopic) BO iteration starting from a sample point under consideration.
Rollout acquisition functions represent state-of-the-art in BO and are 
integrals over $h$ dimensions, where the integrand itself is evaluated through inner optimizations, 
resulting in an expensive integral
that is typically evaluated by Monte Carlo methods.
%
%
The rollout acquisition function is then maximized to determine 
the next BO evaluation, further increasing the cost. This large computational overhead has been 
observed by Osborn et al.~\cite{Osborne2009}, who are only able to compute rollout acquisition for 
horizon 2, dimension 1. Lam et al.~\cite{Lam2016}, who use Gauss-Hermite quadrature in horizons 
up to five, saw runtimes on the order of hours for small, synthetic functions \cite{Frazier2019}.

Recent work focuses on making rollout more practical. Wu and Frazier \cite{Frazier2019} consider horizon 2, using a 
combination of Gauss-Hermite quadrature \cite{liupierce1994} and Monte Carlo (MC) integration to quickly calculate the 
acquisition function and its gradient. Non-myopic active learning also uses rollout \cite{garnett2012bayesian, jiang2017efficient, jiang2018efficient, krause2007nonmyopic}
and recent work develops a fast implementation by truncating the horizon and selecting a 
batch of points to collect future rewards~\cite{jiang2017efficient, jiang2018efficient}.

\subsection{Gaussian process regression}
A Gaussian process (GP) is a collection of random variables, any finite number of which have a joint Gaussian distribution. Gaussian processes can be used to describe a distribution over functions.
We place a GP prior on $f(\bfx)$, denoted by $f \sim \GP(\mu, k)$, where 
$\mu : \Omega \to \bbR$ and $k : \Omega \times \Omega \to \bbR$ are the mean and 
covariance function, respectively. Here,  $k$ is a kernel that correlates points in our sample space and it 
typically contains hyperparameters—like a lengthscale factor—that are learned to improve the quality
of the approximation \cite{rasmussen_i._2006}.

After observing $f$ at data points $\calD_n = \{(\bfx^i, y_i) : 0 \leq i \leq n, i \in \bbN\}$, the posterior distribution on $f$ is a GP.
We assume the observations are $y_i = f(\bfx^i) + \epsilon_i$ where the $\epsilon_i$ are independent variables
distributed as $\calN(0, \sigma^2)$. Given $\calD_n$, we define the following for convenience of representation:
\[
  \bfy = \begin{bmatrix} y_1\\ \vdots \\ y_n\end{bmatrix}, 
  \quad
  \mu_X = \begin{bmatrix}
  \mu(\bfx^1) \\ \vdots \\ \mu(\bfx^n)
  \end{bmatrix},
  \quad
  \bfk(\bfx) =  \begin{bmatrix}k(\bfx, \bfx^1) \\ \vdots \\ k(\bfx, \bfx^n) \end{bmatrix}, \quad 
  K = \begin{bmatrix}\bfk(\bfx^1)^T \\ \vdots \\  \bfk(\bfx^n)^T\end{bmatrix}.
\]

The resulting posterior distribution for function values at a location $\bfx$ is
the normal distribution $\mathcal{N}(\mu^{(n)}(\bfx\;|\;\calD_n),\; 
K^{(n)}(\bfx,\bfx \; | \;\calD_n))$:
\begin{align}
  \mu^{(n)}(\bfx\;|\;\calD_n) &= 
    \mu(\bfx) + \bfk(\bfx)^T(K+\sigma^2I_n)^{-1}(\bfy - \mu_X) \\
  K^{(n)}(\bfx,\bfx\;|\;\calD_n) &= 
    k(\bfx, \bfx) - \bfk(\bfx)^T(K+\sigma^2I_n)^{-1}\bfk(\bfx)
\end{align}
where $I_n \in \bbR^{n\times n}$ is the identity matrix. GPs allow us to perform inference while quantifying our uncertainty. They also provide us with a mechanism for sampling
possible belief states.

\subsection{Myopic Bayesian optimization}
Consider the problem of seeking a global minimum of a continuous objective $f(\bfx)$ over a 
compact set $\Omega \subseteq \bbR^d$. If $f(\bfx)$ is expensive to evaluate, then 
finding a minimum should be sample-efficient. Myopic Bayesian Optimization (MBO) typically uses a GP to model 
$f(\bfx)$ from the data $\calD_n$. The next evaluation location $\bfx^{n+1}$ is determined by maximizing an 
acquisition function $\alpha(\bfx \;|\; \calD_n)$ where the next sample is made
based on a short-term view, looking only one step ahead:
\[
  \bfx^{n+1} = \arg\max_{\Omega} \alpha(\bfx \;|\; \calD_n).
\]
In the myopic setting, our sequential decision-making problem makes choices that depend on all past
observations, optimizing for some immediate reward. However, we usually have more than one step in our
budget, so myopic acquisition functions are leaving something behind.

\subsection{Non-myopic Bayesian optimization}
Non-myopic BO frames the exploration-exploitation problem as a balance of immediate and future 
rewards. Lam et al.~\cite{Lam2016} formulate non-myopic BO as a finite horizon dynamic program; the equivalent 
Markov decision process follows.

The notation used is standard (see Puterman~\cite{puterman2014markov}): an MDP is a 
collection $(T, \mathbb{S}, \mathbb{A}, P,  R)$, where $T = \{1,2,\dots,h\},$ 
and $h < \infty$ is the set of decision epochs, 
finite for our problem.
The state space, $\mathbb{S}$, encapsulates 
the information needed to model the system 
from time $t \in T$, and $\mathbb{A}$ is the action space.  Given a state $s \in \mathbb{S}$ and an 
action $a \in \mathbb{A}$, $P(s'|s,a)$ is the probability the next state will be $s'$. 
$R(s,a,s')$ is the reward received for choosing action $a$ in state $s$
and transitioning to $s'$.

A decision rule $\pi_t : \mathbb{S} \to \mathbb{A}$ maps states to actions at time $t$. A policy 
$\pi$ is a series of decision rules $\pi = (\pi_1, \pi_2, \dots, \pi_{h})$, one at each decision 
epoch. Given a policy $\pi$, a starting state $s_0$, and horizon $h$, we can define the expected 
total reward $V_h^\pi(s_0)$ as:
\[
  V_h^\pi(s_0) = \bbE\left[ \sum_{t=0}^{h-1} R(s_t, \pi_t(s_t), s_{t+1}) \right].
\]

Our objective is to find the optimal policy 
$\pi^*$ that maximizes the expected total reward, i.e., $\sup_{\pi \in \Pi}V_h^\pi(s_0)$, where $\Pi$ 
is the space of all admissible policies.

If we can sample from the transition probability $P$, we can estimate the expected total reward of
any base policy --- the decisions made using the base acquisition function --- with MC integration ~\cite{Sutton1998}:
%
%
\[
  V_h^{\hat{\pi}}(s_0) \approx 
  \frac{1}{N}\sum_{i=1}^N\left[\sum_{t=0}^{h-1}R(s_t^i, \hat{\pi}_t(s_t^i), s^i_{t+1})\right].
\]

Given a GP prior over data $\calD_t$ with mean $\mu^{(t)}$ and covariance matrix $K^{(t)}$, we 
model $h$ steps of BO as an MDP. This MDP’s state space is all possible data sets reachable from 
starting-state $\calD_t$ with $h$ steps of BO. Its action space is $\Omega$; actions 
correspond to sampling a point in $\Omega$. Its transition probability and reward function are 
defined as follows. Given an action $x^{t+1}$, the transition probability from $\calD_t$ to 
$\calD_{t+1}$, where $\calD_{t+1} = \calD_t \;\cup\; \{(\bfx^{t+1}, 
y_{t+1})\}$ is:
\[
  P(\calD_t, \bfx^{t+1}, \calD_{t+1}) \sim 
    \mathcal{N}(\mu^{(t)}(\bfx^{t+1};\calD_t),K^{(t)}(\bfx^{t+1}, \bfx^{t+1};\calD_t)).
\]
Thus, the transition probability from $\calD_t$ to $\calD_{t+1}$ is the probability of 
sampling $y_{t+1}$ from the posterior $\mathcal{GP}(\mu^{(t)}, K^{(t)})$ at $\bfx^{t+1}$. We 
define a reward according to expected improvement (EI) ~\cite{Jones1998}. Let $f^*_t$ be 
the minimum observed value in the observed set $\calD_t$, i.e., $f^*_t = \min\{y_0, \dots, 
y_t\}$. Then our reward is expressed as follows:
\[
  R(\calD_t, \bfx^{t+1}, \calD_{t+1}) = (f^*_t-f_{t+1})^+ \equiv \max(f^*_t-f_{t+1}, 0).
\]
EI can be defined as the optimal policy for horizon one, obtained by maximizing the immediate reward:
\[
  \pi_{EI} 
  = \argmax_\pi V_1^\pi(\calD_t) 
  = \argmax_{\bfx^{t+1} \in \Omega} \bbE\left[ (f_t^*-f_{t+1})^+ \right] 
  \equiv \argmax_{\bfx^{t+1}\in\Omega} EI(\bfx^{t+1}|\calD_t),
\]
where the starting state is $\calD_t$—our initial samples. 
We define the non-myopic policy as the optimal solution to an $h$-horizon MDP. The expected total reward of this MDP is:
\[
  V_h^\pi(\calD_n) = \bbE\left[ \sum_{t=n}^{n+h-1} R(\calD_t, \pi_t(\calD_t), \quad
  \calD_{t+1})\right] = \bbE\left[ \sum_{t=n}^{n+h-1} (f_t^* - f_{t+1})^+\right] =
    \bbE\left[ \left(f_n^* - \min_{t \leq n+h}f_{t}\right)^+\right].
\]
The integrand is a telescoping sum, resulting in the equivalent expression on the rightmost side of the equation.
When $h>1$, the optimal policy is difficult to compute.

\subsection{Rollout acquisition functions}
Finding an optimal policy for the $h$-horizon MDP is hard, partly because of the expense of representing and optimizing over policy functions.
One approach to faster approximate solutions to the MDP is to replace the infinite-dimensional space of policy functions with a family of trial
policies described by a modest number of parameters.
An example of this approach is {\em rollout} policies~\cite{Bert05}, which yield promising results for the $h$-step horizon MDP~\cite{Frazier2019}.
For a given state $\calD_n$, we say that a base policy $\Tilde{\pi} $ is a sequence of rules that determines the actions
to be taken in various states of the system. 

We denote  our base policy for the $h$-horizon MDP as $\Tilde{\pi} = (\Tilde{\pi}_0, \Tilde{\pi}_1, \Tilde{\pi}_2, \dots, \Tilde{\pi}_h)$. We let
$\calD_n$ denote the initial state of our MDP and $\calD_{n,t}$ for $0 \leq t \leq h$ 
to denote the random variable that is the state at each decision epoch. Each individual decision rule
$\Tilde{\pi}_t$ 
for $t \geq 1$ 
consists of maximizing the base acquisition function $\bar{\alpha}$ given the current
state $s_t = \calD_{n, t}$, i.e.
\[
  \Tilde{\pi}_t = \argmax_{\bfx \in \Omega} \bar{\alpha} \left( \bfx | \calD_{n,t}\right).
\]
Using this policy, we define the non-myopic acquisition function $\alpha_h(\bfx)$ as
the rollout of $\Tilde{\pi}$ to horizon $h$, i.e. the expected reward of $\Tilde{\pi}$ starting with action
$\Tilde{\pi}_0 = \xstart$:
\[
  \alpha_h\left( \xstart \right) 
  \coloneqq 
  \bbE \left[ V^{\Tilde{\pi}}_{h} \left( \calD_{n} \cup \{ \left(\xstart, \ystart\right) \} \right) \right],
\]
where $\ystart$ is the noisy observed value of $f$ at $\xstart$. Thus, as is the case with
any acquisition function, the next BO evaluation is:
\[
  \xnext = \argmax_{\xstart \in \;\Omega} \alpha_h\left(\xstart\right).
\]
Rollout is tractable and conceptually straightforward, however, it is still computationally demanding.
To rollout $\Tilde{\pi}$ once, we must do $h$ steps of BO with $\bar{\alpha}$. Many of the
aforementioned rollouts must then be averaged to reasonably estimate $\alpha_h$, which is an
$h$-dimensional integral. Estimation can be done either through explicit quadrature or MC integration,
and is the primary computational bottleneck of rollout.

\section{Models and methods}
\label{sec:models}

\label{models_and_methods}
\subsection{Model}
We build our intuition behind our approach from a top-down perspective. We have seen that non-myopic 
Bayesian optimization is promising, though tends to be computationally intractable. To circumvent
this problem, we formulate a sub-optimal approximation to solve the intractable dynamic program;
namely, a rollout acquisition function. Though relatively more tractable, rollout acquisition
functions can be computationally burdensome. We are interested in solving $x^{*} = \argmax_{x \in 
\mathcal{X}} \alpha_h(x)$ where
\begin{equation}
        \alpha_h(\bfx) = \bbE_{\hat{f}^{i}\sim\mathcal{G}}
        \left[\alpha\big(\bfx|\tau(h,\bfx,\bar{\alpha},\hat{f}^{i})\big)\right]
        \approx \frac{1}{N}\sum_{i=1}^N (f^*-\bigl(t^i(\bfx)\bigr)^-)^+
\end{equation}
and $t^i(x) \sim \tau(h,x,\bar{\alpha},\hat{f}^{i})$ are sample trajectories. Sample trajectories
consist of picking a starting point and asking ``what would happen if I used my base policy from this
point forward?'' Since these trajectories don't represent what did happen, rather what might, we call
each realization a fantasized sample.
Unfortunately, derivative-free optimization in high-dimensional spaces is expensive, so
we would like estimates of $\nabla_{x}\alpha_h(x)$. In particular, differentiating with respect to $x$
yields the following:
\begin{equation}
    \nabla_x\left[\alpha_h(\bfx)\right] \approx \nabla_x
    \left[\frac{1}{N}\sum_{i=1}^N (f^*-\bigl(t^i(\bfx)\bigr)^-)^+\right]
    = \frac{1}{N}\sum_{i=1}^N \left(-\nabla_x\left[\bigl(t^i(\bfx)\bigr)^-\right]\right)^+
\end{equation}
which requires some notion of differentiating sample trajectories. Thus, our problem can be expressed as
two interrelated optimizations:
\begin{enumerate}
    \item An inner optimization for determining sample trajectories $t^i(\bfx) \sim \tau(h, \bfx, \bar{\alpha}, \hat{f}^{i})$
    \item An outer optimization for determining our next query location $\bfx^*$
\end{enumerate}
In what follows, we define how to compute and differentiate sample trajectories and how to differentiate the 
rollout acquisition function. We introduce the following notation to distinguish between two GP sequences that must be maintained. Suppose
we have function values and gradients denoted as follows: $(\hat{f}_0, \nabla\hat{f}_0), \dots
(\hat{f}_h, \nabla\hat{f}_h)$. We define two GP sequences as:
\begin{align*}
    \mathcal{F}_h \; \sim \;\mathcal{F}_0 &\; | \;\mathcal{F}_h(\bfx^0)=\hat{f}_0,\dots,\mathcal{F}_h(\bfx^h)=\hat{f}_h \\
    \mathcal{G}_h \; \sim \;\mathcal{F}_0 &\; | \;\mathcal{F}_h(\bfx^0)=\hat{f}_0,\dots,\mathcal{F}_h(\bfx^h)=\hat{f}_h, \\
    &\nabla\mathcal{F}_h(\bfx^0)=\nabla\hat{f}_0,\dots,\nabla\mathcal{F}_h(\bfx^h)=\nabla\hat{f}_h.
\end{align*}

The model, fundamentally, relies 
on the distinction amongst known sample locations $\{\bfx^{-j} \in\bbR^d \;|\; 1 \leq j \leq 
m\}$, deterministic start location $\{\bfx^0 \in\bbR^d\}$, and the stochastic fantasized 
sample locations $\{\bfx^j \in\bbR^d\;|\; 1 \leq j \leq h\}$. We denote a full trajectory including
$h$ fantasized steps by:
\[
X^{m+h+1} :=
\begin{bmatrix} \bfx^{-m} \;\dots \;\bfx^{-1} | \;\bfx^0 | \; \bfx^1 \;\dots\; \bfx^h\end{bmatrix}
\in \bbR^{d\times (m+h+1)}.
\]
Moreover, we collect the $m$ known samples into $y=\begin{bmatrix}y_{-m}\; \dots 
\;y_{-1}\end{bmatrix}^T$. 
The distinction between negative, null, and positive superscripts serves a useful purpose. Negative superscripts can be thought of as the past; potentially noisy observations that are fixed and immutable.
The null superscripts denotes our freedom of choice; we are not bound to start at a specific location.
Positive superscripts can be thought of as our ability to see into the future: things we anticipate on observing given our current beliefs.

Our problem is how to choose $\bfx^0$ to maximize our expected reward over some finite horizon $h$. We focus on developing an $h$-step expected improvement (EI) rollout policy that involves choosing $\bfx^0$ by using Stochastic Gradient Ascent (SGA) to optimize our rollout acquisition function.

An $h$-step EI \textit{rollout} policy involves choosing $\bfx^0$ based on the anticipated behavior of 
the EI ($\bar{\alpha}$) algorithm starting from $\bfx^0$ and proceeding for $h$ steps. That is, we consider the 
iteration
\begin{equation}
    \bfx^r = \argmax_x\bar\alpha(\bfx \;|\; \mathcal{F}_{r-1}), \; 1 \leq r \leq h
\end{equation}
where the trajectory relations are
\begin{align*}
    \nabla_{x}\bar{\alpha}(\bfx^r|\mathcal{F}_{r-1})=\textbf{0} \;\land\; \left(\hat{f}_r, 
    \nabla\hat{f}_r\right) \sim \mathcal{G}_{r-1}(\bfx^r).
\end{align*}

Trajectories are fundamentally random variables; their 
behavior is determined by the rollout horizon $h$, start location $\bfx$, base policy $\bar{\alpha}$, and 
initial observation $\hat{f}^i$—denoted as $\tau(h,\bfx, \bar{\alpha}, \hat{f}^i)$. 
We denote sample draws $t^i \sim \tau(h, \bfx, \bar{\alpha}, \hat{f}^i)$ as follows:
%
\begin{equation}
    t^i = \bigg( \Big(\bfx^j, \hat{f}_j^i(\bfx^j), \nabla\hat{f}_j^i(\bfx^j)\Big) \bigg)_{j=0}^h  
\end{equation}

\begin{figure}[ht]
    \centering
    \subfigure{\includegraphics[width=0.49\textwidth]{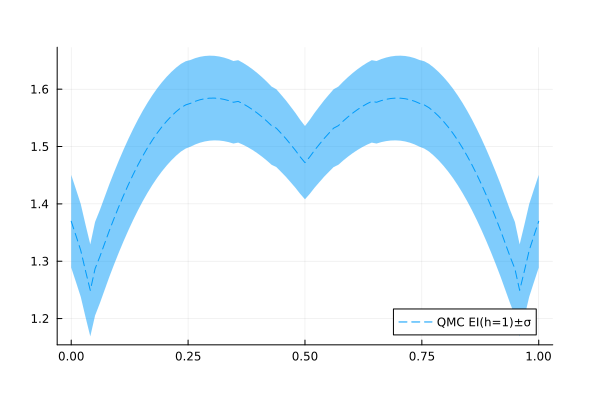}}
    \subfigure{\includegraphics[width=0.49\textwidth]{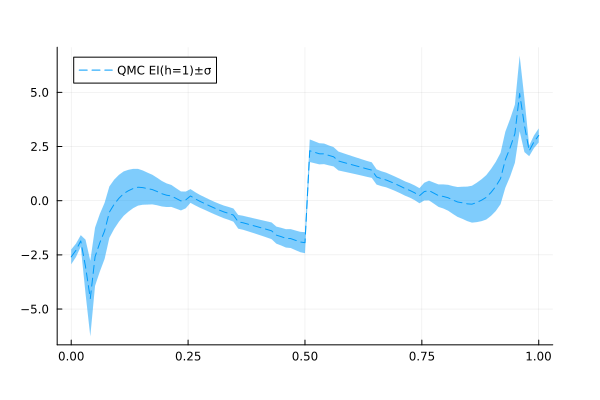}}
    
    \caption{The first graph is the rollout acquisition function for a non-trivial horizon. Subsequently,
    we depict the estimated gradient and their respective standard errors.}
    \label{fig:myfigure}
\end{figure}

Computing this sample path, however, requires we solve the iteration defined above. We also use the notation $t^i_{j,k}$ 
to denote the $k$-th element of the $j$-th 3-tuple associated with the $i$-th sample.
Now that we have some provisional notation, we define how to evaluate $t^i(x)$ as follows:
\begin{equation}
    (t^i(\bfx))^- \equiv \min(t^i(\bfx)) \equiv \min_{0\leq j\leq h}t^i_{j,2}
\end{equation}
If we let $b = \argmin_j t^i_{j,2}$, we can rewrite the minimum of the sample trajectory as follows:
\begin{equation}
    (t^i(\bfx))^- \equiv \hat{f}^i_b(\bfx^b(\bfx)).
\end{equation}
Now, we are able to differentiate trajectories given by the following
\begin{equation}
    \nabla_x\left[(t^i(\bfx))^-\right] = \begin{bmatrix}\frac{\partial}{\partial x_1}\left[(t^i(\bfx))^-\right],\;\dots,\;
    \frac{\partial}{\partial x_d}\left[(t^i(\bfx))^-\right]\end{bmatrix}^T
\end{equation}
which requires us to compute $\frac{\partial}{\partial x_k}\left[(t^i(\bfx))^-\right]$ where $ 1\leq k \leq d$. Hence,
we have
\begin{equation}
    \frac{\partial}{\partial x_k}\left[(t^i(\bfx))^-\right] = \frac{\partial}{\partial x_k}\left[\hat{f}_b(\bfx^b(\bfx))\right].
\end{equation}
Since the best function value $\hat{f}_b$ found is a function of the best location $x^b$, which is also a function of
$x$, we have
\begin{equation}
    \frac{\partial}{\partial x_k}\left[\hat{f}_b(\bfx^b(\bfx))\right] = \frac{\partial\hat{f}_b}{\partial \bfx^b}\frac{\partial \bfx^b}
    {\partial x_k}.
\end{equation}
This captures how the best value found thus far changes as we vary the $k$-th dimension of $x$ which is equivalently
expressed as
\begin{equation}
    \nabla_x\left[\hat{f}_b(\bfx^b(\bfx))\right] = \left(\hat{f}_b'(\bfx^b(\bfx))\cdot\frac{\partial \bfx^b}{\partial x}\right)^T.
\end{equation}
From the trajectory relations defined above, we are able to derive the following relationship via the Implicit Function
Theorem about $x^r$:
\begin{equation*}
    \frac{\partial\bar{\alpha}}{\partial \bfx^r}\left(\bfx^r|\mathcal{F}_{r-1}\right)=\textbf{0} \rightarrow\\ 
    \frac{\partial^2\bar{\alpha}}{(\partial \bfx^r)^2}\frac{\partial \bfx^r}{\partial \bfx} +\frac{\partial^2\bar{\alpha}}
    {\partial \bfx^r \partial\mathcal{F}_{r-1}}\frac{\partial\mathcal{F}_{r-1}}{\partial \bfx} = \textbf{0}.
\end{equation*}
We are able to compute the Jacobian matrix $J(\bfx^b) = \frac{\partial \bfx^b}{\partial \bfx}$ as
\begin{equation}
    \frac{\partial \bfx^b}{\partial \bfx} = -\left(\frac{\partial^2\bar{\alpha}}{(\partial \bfx^b)^2}\right)^{-1}
    \left(
    \frac{\partial^2\bar{\alpha}}{\partial \bfx^b \partial\mathcal{F}_{b-1}}
    \frac{\partial\mathcal{F}_{b-1}}{\partial \bfx}
    \right)
\end{equation}
defining all of the computations necessary to solve the inner and outer optimization problem.

\subsection{Methods}
\begin{figure}[ht]
    \centering
    \subfigure{\includegraphics[width=0.49\textwidth]{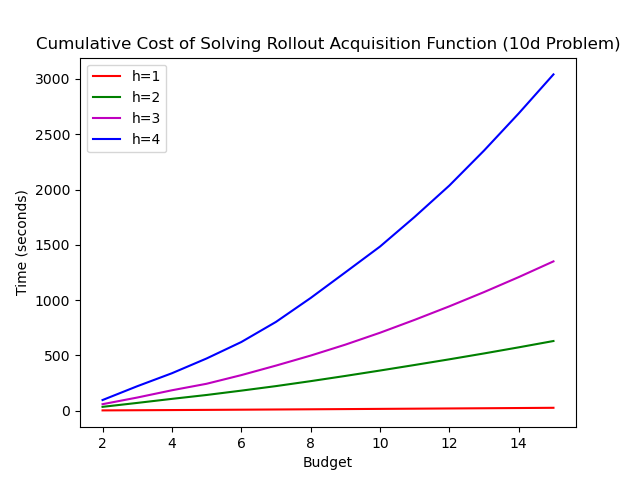}}
    \subfigure{\includegraphics[width=0.49\textwidth]{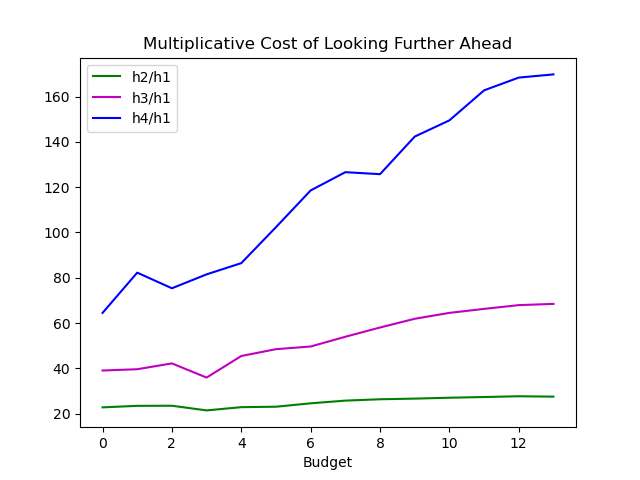}}
    
    \caption{The first graph demonstrates the cost of 
    increasing your horizon as your iterations progress. 
    Subsequently,
    we depict the multiplicative cost of looking further 
    ahead relative to the rollout acquisition function with 
    $h=1$. Note, $h=1$ corresponds to a two-step lookahead strategy.}
    \label{fig:myfigure}
\end{figure}

Significant computational challenges arise when computing the rollout acquisition function (RAF). Namely, rolling out the base acquisition 
function for $h$ 
steps consists of updating and maintaining a fantasized GP which is averaged across some $N$ monte-carlo simulations. Naive updates of the 
covariance and Cholesky matrix for each fantasized GP unnecessarily repeats computations. Another cost incurred when evaluating the rollout 
acquisition function (RAF) is that the $h$-step iteration is solving an optimization problem at each step. Finally, in order to differentiate the 
RAF, we need to differentiate trajectories, requiring us to estimate a sequence of Jacobian matrices. In what follows, we address the following
techniques to enable efficient computation and optimization of the RAF: smart linear algebra, quasi-monte carlo, common random numbers and
control variates.

\textbf{Differentiation of the base acquisition function}
Evaluating the rollout acquisition function necessitates the optimization of the base acquisiton function iteratively, for a total of $h$ times.
To expedite this step, our approach uses the gradient and 
Hessian of the base policy. This
enables us to use second-order optimization algorithms that 
significantly speeds up the optimization process.

\textbf{Smart linear algebra.}
Maintaining $N$ GPs with an $h$-step base policy for some evaluation of $\alpha_h$ at location $x^0$ is expensive. The value of $h$ determines
how many inner optimization problems we solve and we call $\alpha_h(x^0)$ the $(h+1)$-step lookahead acquisition function. Computing a
covariance matrix and it's inverse is $\mathcal{O}(n^2)$ and $\mathcal{O}(n^3)$, respectively. Doing this for our $h$-step policy produces
something on the order of $\sum_{i=0}^h \mathcal{O}((n + i)^2)$ for the covariance updates and $\sum_{i=0}^h \mathcal{O}((n + i)^3)$
for the matrix inversion updates. To circumvent this inefficiency, we preallocate matrices for both the covariance and Cholesky factorizations. 
This preallocation is dynamically updated using Schur’s complement. This approach yields a substantial reduction in computational load by making 
updates something on the  order of a polynomial of a lower integer degree, thereby streamlining the inner optimization process.

\subsubsection{Variance reduction techniques}
Variance reduction is a set of methods that improve convergence by decreasing the variance of some estimator.
Useful variance reduction methods can reduce the sample variance by several orders of magnitude~\cite{Opc1993}. We use a combination of
quasi-Monte Carlo, common random numbers, and control variates, which significantly reduces the number of MC samples needed, while smoothing
out estimates of the underlying stochastic objective.

\textbf{Quasi-monte carlo. (QMC)} Monte Carlo methods use independent, uniformly distributed random numbers on the $p$-dimensional unit cube as
the source of points to integrate at. The distribution we integrate over is Normal, for which a low-discrepancy sequence is known to exist. We 
generate low-discrepancy Sobolev sequences in the $p$-dimensional uniform distribution $\mathcal{U}[0, 1]^p$ and map them to the standard 
multivariate Gaussian via the Box-Muller transform~\cite{Society1992}. This produces a low-discrepancy sequence for $\mathcal{N}(0, I_p)$, 
allowing us to apply QMC by replacing the standard multivariate Gaussian samples with our low-discrepancy sequence.

\textbf{Common random numbers (CRN).} In many stochastic processes, different iterations or simulations use separate sets of random numbers to 
generate outcomes. While this approach is robust in ensuring the independence of each trial, it can also introduce a high degree of variability 
or noise into the results. This is where Common Random Numbers come into play. The CRN techniques involves using the same set of random
numbers across different iterations. CRN does not decrease the pointwise variance of an estimate, but rather decreases the covariance between 
two  neighboring estimates, which smooths out the function.

\textbf{Control variates.} A control variate is a variable that is highly correlated with the 
function of interest and has a known expected value. The key is to find a variable whose 
behavior is similar to that of the function being estimated. 
Since $\alpha_h(x) = 
\mathbb{E}\left[ \hat{\alpha}_h(x) \right]$ corresponds
to an $h$-step EI acquisition, we use the $1$-step EI acquisition as our control variate.
That is, we can estimate $\alpha_h^{\operatorname{cv}}(x) = \mathbb{E}\left[ \hat{\alpha}_h(x) + \beta w(x) \right]$ where $w(x) = \operatorname{max}(f^+ - f(x), 0) - \operatorname{EI}(x)$. Given this construction, the variance of our estimator is given as follows:
\begin{equation*}
\begin{split}
    \mathrm{Var}\left(\alpha_h^{\operatorname{cv}}(x)\right) &= \mathrm{Var}\left(\hat{\alpha}_h(x) + \beta w(x)\right) \\
    &= \mathrm{Var}\left(\hat{\alpha}_h(x)\right) + \beta^2 \mathrm{Var}\left(w(x)\right) + 2\beta\mathrm{Cov}\left(\hat{\alpha}_h(x), w(x)\right).
\end{split}
\end{equation*}
The variance of our estimator is quadratic in $\beta$ and is minimized for
$\beta^{*} = \frac{-\mathrm{Cov(\hat{\alpha}, w)}}{\mathrm{Var}(w)}$.  However, since $\mathrm{Cov(\hat{\alpha}, w)}$ is unknown, we estimate $\beta^*$ from the Monte Carlo simulations.

Overall, by combining smart linear algebra, quasi-Monte Carlo integration, common random numbers, and control variates, we show that computing
and differentiating rollout acquisition functions is tractable.

\section{Experiments and discussion}
\label{sec:experiments}
\begin{figure}[htbp]
    \centering
    \includegraphics[width=\textwidth]{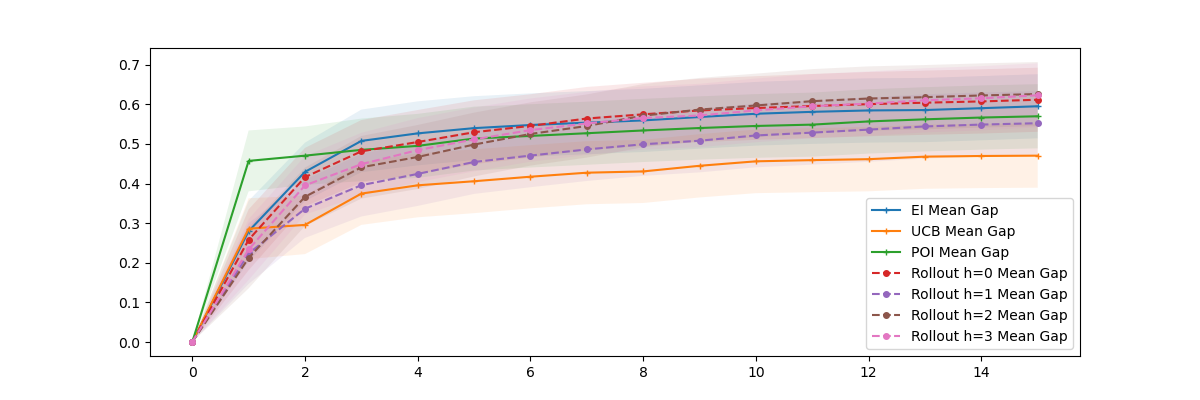}
    \caption{The average GAP and standard error across a suite of 15 synthetic test functions with 60 randomized trials and a single 
    random observation as an initialization of the underlying GP model. All nonmyopic strategies outperform POI and most outperform
    EI.}
    \label{fig:example}
\end{figure}
Throughout our experiments we use a GP with the Matérn 5/2 ARD kernel \cite{Snoek2012a} and learn its hyperparameters via maximum likelihood estimation ~\cite{rasmussen2003gaussian}.
When rolling out acquisition functions, we maximize them using the Adam-variant of Stochastic Gradient Ascent (SGA) with expected improvement used as the base policy. We use the proposed default values for Adam of 0.9 for $\beta_1$, $0.999$ for $\beta_2$, and $10^{-8}$ for $\epsilon$~\cite{kingma2017adam}. We use the evidence-based criterion developed in~\cite{Mahsereci2017} as our convergence criteria.
We use $8$ random restarts for solving the inner optimization with a maximum of
$50$ iterations for SGA and select the best point found. All synthetic functions are found in the supplementary.

Given a fixed evaluation budget, we evaluate the performance of an algorithm in terms of its gap $G$. The gap measures the best decrease in the objective function from the first to the last iteration, normalized by the maximum reduction possible:
$$
G = \frac{
f_{\operatorname{min}}^{\mathcal{D}_1} - f_{\operatorname{min}}^{\mathcal{D}_{B+1}}
}{
f_{\operatorname{min}}^{\mathcal{D}_1} - f(x^*)
}
$$

\begin{figure}[h]
    \centering
    \includegraphics[width=0.95\textwidth]{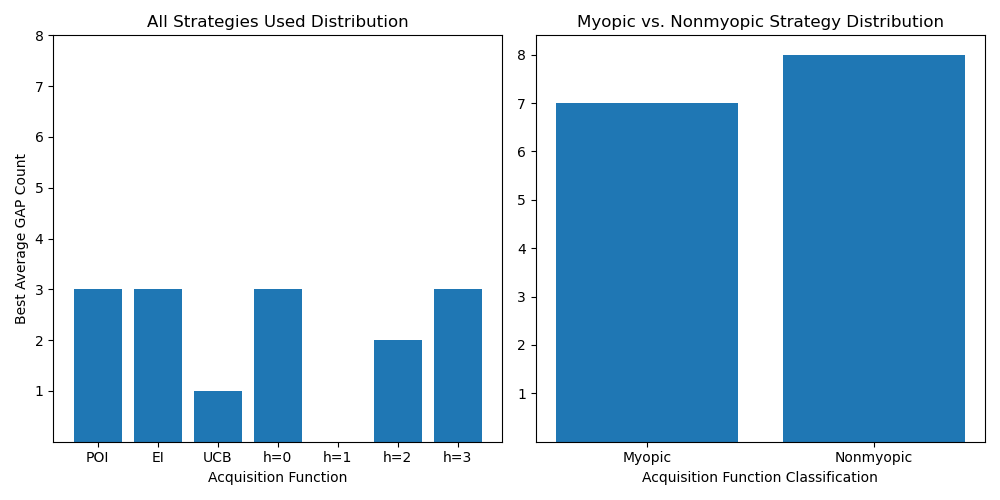}
    \caption{A histogram depicting the average winning strategies.}
    \label{fig:your_label}
\end{figure}


\begin{table}[ht]
\centering
\caption{Mean and median values for various functions and methods ran
for 15 iterations of BO over 60 initial guesses.}
\label{my-label}
\begin{tabularx}{\textwidth}{@{}l *{9}{X}@{}}
\toprule
Function name & & PI & EI & UCB & KG & \(\alpha_0\) & \(\alpha_1\) & \(\alpha_2\) & \(\alpha_3\) \\ \midrule
Gramacylee          & Mean   & .590 & .594 & .268 & .608 & .750 & .665 & .842 & \textbf{.875} \\
                    & Median & .585 & .631 & .181 & .700 & .796 & .707 & \textbf{.997} & .991\\
Schwefel4d          & Mean   & .074 & .075 & .124 & \textbf{.257} & .073 & .140 & .137 & .163 \\
                    & Median & .015 & .017 & .017 & \textbf{.262} & 0.0 & .124 & .090 & .057 \\
Rosenbrock          & Mean   & .606 & .435 & .425 & \textbf{.874} & .813 & .683 & .594 & .597 \\
                    & Median & .806 & .431 & .403 & \textbf{.975} & .954 & .885 & .845 & .847 \\
Branin-Hoo          & Mean   & .440 & .533 & .318 & \textbf{.888} & .506 & .507 & .550 & .490 \\
                    & Median & .438 & .582 & 0.0  & \textbf{.947} & .576 & .623 & .695 & .598 \\
Goldstein-Price     & Mean   & .770 & .712 & .099 & \textbf{.928} & .818 & .784 & .770 & .742 \\
                    & Median & .932 & .877 & 0.0  & \textbf{.983} & .959 & .892 & .939 & .935 \\
Six-Hump Camel      & Mean   & .808 & .726 & .279 & .823 & .799 & .768 & .910 & \textbf{.934} \\
                    & Median & .892 & .861 & 0.0  & .934 & .917 & .874 & .961 & \textbf{.965} \\
\bottomrule
\end{tabularx}
\end{table}

\section{Limitations}
\label{sec:limitations}
The number of experiments conducted was limited due to hardware
considerations. The experiments were run on a Intel(R) Core(TM) 
i7-7700 CPU @ 3.60GHz Linux workstation with 8 CPU cores.
We did not treat some important considerations that might yield 
better results. Namely, our MDP model does not consider a discount 
factor,
which may yield different behaviors. When solving both the myopic and 
non-myopic acquisition functions, our tolerance in 
$\bfx^{\operatorname{opt}}$ is $1 
\text{e}{-3}$. Using the implicit function theorem, we estimate 
the sequence of Jacobians using forward-mode 
differentiation, introducing a non-trivial expense. 

\section{Conclusion}
\label{sec:conclusion}
Our insights yield several interesting research directions. We have shown how to construct and optimize over a parametric family of
viable policy functions, also known as a policy gradient method, using well-known myopic acquisition functions. This formulation also
enables the treatment of the horizon as an adaptive hyperparameter learned as one iterates. Furthermore, when comparing one-step against
multi-step lookahead strategies using the same base acquisition function, we observe that there is no definitive best strategy. This suggest
that a finite horizon should not be fixed for the entire iteration.

Our contributions focused on using expected improvement as our
base policy, but our analysis is base policy agnostic and only
assumes the given base policy has two derivatives.


\small
\nocite{*}
\bibliography{refs}

\section*{Appendix}


\appendix

\section{Kernels}
The kernel functions used in this paper is the Matérn 5/2 kernel:
\begin{equation*}
    K_{5/2}(r) = \sigma^2\left(1 + \frac{\sqrt{5}}{\ell} + \frac{5}{3\ell^2}\right)
    \exp(-\frac{\sqrt{5}r}{\ell})
\end{equation*}

\section{Expected Improvement}
\begin{align*}
    \alpha(x) &= \sigma(x) g(z(x)) \\
    g(z) &= z \Phi(z) + \phi(z) \\
    z(x) &= \sigma(x)^{-1} \left[ \mu(x) - f^+ - \xi \right]  
\end{align*}

\section{Differentiating $\alpha$}
Differentiating with respect to spatial coordinates:
\begin{align*}
  \alpha_{,i} &= \sigma_{,i} g(z) + \sigma g'(z) z_{,i} \\
  \alpha_{,ij} &= \sigma_{,ij} g(z) + \sigma_{,i} g'(z) z_{,j} + \sigma_{,j} g'(z) z_{,i} + \sigma g'(z) z_{,ij} + \sigma g''(z) z_{,i} z_{,j} \\
  &= \sigma_{,ij} g(z) + [\sigma_{,i} z_{,j} + \sigma_{,j} z_{,i} + \sigma z_{,ij}] g'(z) + \sigma g''(z) z_{,i} z_{,j}
\end{align*}

Mixed derivative with respect to spatial coordinates and data and hypers:
\begin{align*}
  \dot{\alpha}_{,i} &= \dot{\sigma}_{,i} g(z) + \sigma_{,i} g'(z) \dot{z} + \dot{\sigma} g'(z) z_{,i} + \sigma g''(z) \dot{z} z_{,i} + \sigma g'(z) \dot{z}_{,i}
\end{align*}

Finally, we differentiate $g(z) = z \Phi(z) + \phi(z)$, noting that
$\phi'(z) = -z \phi(z)$ and $\Phi'(z) = \phi(z)$.  This gives
\begin{align*}
  g(z) &= z \Phi(z) + \phi(z) \\
  g'(z) &= \Phi(z) + z \phi(z) + \phi'(z) = \Phi(z) \\
  g''(z) &= \phi(z).
\end{align*}

\section{Differentiating $z$}
Now consider $z = \sigma^{-1} [\mu - f^+ - \xi]$.  As before, we begin with spatial derivatives:
\begin{align*}
  z_{,i} &= -\sigma^{-2} \sigma_{,i} [\mu - f^+ - \xi] + \sigma^{-1} \mu_{,i} \\
         &= \sigma^{-1} \left[ \mu_{,i} - \sigma_{,i} z \right] \\
  z_{,ij} &= -\sigma^{-2} \sigma_{,j} \left[ \mu_{,i} - \sigma_{,i} z \right] + 
             \sigma^{-1} \left[\mu_{,ij} - \sigma_{,ij} z - \sigma_{,i} z_{,j} \right] \\
          &= \sigma^{-1} \left[ \mu_{,ij} - \sigma_{,ij} z - \sigma_{,i} z_{,j} - \sigma_{,j} z_{,i} \right]
\end{align*}

Now we differentiate with respect to data and hypers:
\begin{align*}
  \dot{z} &= -\sigma^{-2} \dot{\sigma} [\mu - f^+ - \xi] + \sigma^{-1} [\dot{\mu} - \dot{f}^+ - \dot{\xi}] \\
          &= \sigma^{-1} [\dot{\mu} - \dot{f}^+ - \dot{\xi} - \dot{\sigma} z] \\
  \dot{z}_{,i} &= \sigma^{-1} [\dot{\mu}_{,i} -\dot{\sigma}_{,i} z - \dot{\sigma} z_{,i} -\sigma_{,i} \dot{z}]
\end{align*}

\section{Differentiating $\sigma$}
The predictive variance is
$$
  \sigma^2 = k_{xx} - k_{xX} K_{XX}^{-1} k_{Xx}.
$$
Differentiating the predictive variance twice in space --- assuming $k_{xx}$ is independent of $x$ by stationarity ---
gives us
\begin{align*}
  2 \sigma \sigma_{,i} &= -2 k_{xX,i} K_{XX}^{-1} k_{Xx} = -2 k_{xX,i} d \\
  2 \sigma_{,i} \sigma_{,j} + 2 \sigma \sigma_{,ij} &= -2 k_{xX,ij} K_{XX}^{-1} k_{Xx} - 2 k_{xX,i} K_{XX}^{-1} k_{Xx,j} \\
                     &= -2 k_{xX,ij} d -2 k_{xX,i} w^{(j)}
\end{align*}

Rearranging to get spatial derivatives of $\sigma$ on their own gives us
\begin{align*}
  \sigma_{,i} &= -\sigma^{-1} k_{xX,i} d \\
  \sigma_{,ij} &= -\sigma^{-1} \left[ k_{xX,ij} d + k_{xX,i} w^{(j)} + \sigma_{,i} \sigma_{,j} \right].
\end{align*}

Differentiating with respect to data (and locations) and kernel hypers requires more work.  First, note that
\begin{align*}
  2 \sigma \dot{\sigma} 
  &= \dot{k}_{xx} - 2 \dot{k}_{xX} K_{XX}^{-1} k_{Xx} + k_{xX} K_{XX}^{-1} \dot{K}_{XX} K_{XX}^{-1} k_{Xx} \\
  &= \dot{k}_{xx} - 2 \dot{k}_{xX} d + d^T \dot{K}_{XX} d
\end{align*}

Now, differentiating $\sigma^{-1}$ with respect to data and hypers gives
\begin{align*}
  \dot{\sigma}_{,i} 
  &= \sigma^{-2} \dot{\sigma} k_{xX,i} K_{XX}^{-1} k_{Xx} -
     \sigma^{-1} \left[ 
       \dot{k}_{xX,i} K_{XX}^{-1} k_{Xx} +
       k_{xX,i} K_{XX}^{-1} \dot{k}_{Xx} -
       k_{xX} K_{XX}^{-1} \dot{K}_{XX} K_{XX}^{-1} k_{Xx} \right] \\
  &= -\sigma^{-1} \left[ \dot{\sigma} \sigma_{,i} + \dot{k}_{xX,i} d + (w^{(i)})^T \dot{k}_{Xx} - d^T \dot{K}_{XX} d \right]
\end{align*}

\section{Differentiating $\mu$}
Let $K_{XX}$ denote the kernel matrix, and $k_{Xx}$ the column vector of kernel evaluations at $x$.
The posterior mean function for the GP (assuming a zero-mean prior) is
$$
  \mu = k_{xX} c
$$
where $K_{XX} c = y$.  Note that $c$ does not depend on $x$, but it does depend on the data and hyperparameters.

Differentiating in space is straightforward, as we only invoke the kernel derivatives:
\begin{align*}
  \mu_{,i} &= k_{xX,i} c \\
  \mu_{,ij} &= k_{xX,ij} c
\end{align*}

Differentiating in the data and hyperparameters requires that we also differentiate through a matrix solve:
$$
  \dot{\mu} = \dot{k}_{xX} K_{XX}^{-1} y + k_{xX} K_{XX}^{-1} \dot{y} - k_{xX} K_{XX}^{-1} \dot{K}_{XX} K_{XX}^{-1} y.
$$
Defining $d = K_{XX}^{-1} k_{Xx}$, we have
$$
  \dot{\mu} = \dot{k}_{xX} c + d^T (\dot{y} - \dot{K}_{XX} c).
$$
Now differentiating in space and defining $K_{XX}^{-1} k_{Xx,i}$ as $w^{(i)}$, we have
$$
  \dot{\mu}_{,i} = \dot{k}_{xX,i} c + (w^{(i)})^T (\dot{y} - \dot{K}_{XX} c).
$$

\section{Differentiating kernels}
We assume the kernel has the form $k(x,y) = \psi(\rho)$ where $\rho = \|r\|$ and $r = x-y$.  Recall that
\begin{align*}
  \rho &= \sqrt{r_k r_k} \\
  \rho_{,i} &= \rho^{-1} r_k r_{k,i} = \rho^{-1} r_i \\
  \rho_{,ij} &= \rho^{-1} \delta_{ij} - \rho^{-2} r_i \rho_{,j} \\
             &= \rho^{-1} \left[ \delta_{ij} - \rho^{-2} r_i r_j \right]
\end{align*}

Applying this together with the chain rule yields
\begin{align*}
  k &= \psi(\rho) \\
  k_{,i} &= \psi'(\rho) \rho_{,i} = \psi'(\rho) \rho^{-1} r_i \\
  k_{,ij} &= \psi''(\rho) \rho_{,i} \rho_{,j} + \psi'(\rho) \rho_{,ij} \\
          &= \left[ \psi''(\rho) - \rho^{-1} \psi'(\rho) \right] \rho^{-2} r_i r_j + \rho^{-1} \psi'(\rho) \delta_{ij}.
\end{align*}

\end{document}